\documentclass[runningheads]{llncs}
\usepackage{graphicx}
\usepackage{amsmath,amssymb,xspace,epsfig,color}
\usepackage[linesnumbered,ruled]{algorithm2e}
\let\eps\varepsilon

\long\def\commented#1{}

\let\eps\varepsilon

\begin{document}
\title{Topology Based Scalable Graph Kernels}
\author{Kin Sum Liu\inst{1}\and
Chien-Chun Ni\inst{2} \and
Yu-Yao Lin\inst{3} \and
Jie Gao\inst{1}\vspace{-2mm}}
\institute{Stony Brook Uni., Stony Brook NY 11794, USA \email{\{kiliu,jgao\}@cs.stonybrook.edu} \and
Yahoo Research, Sunnyvale CA 94089, USA \email{cni02@oath.com} \and
Intel Inc., Hillsboro OR 97124, USA \email{yu-yao.lin@intel.com}
}

\maketitle              
\section{Introduction}
We propose a new graph kernel for graph classification and comparison using Ollivier Ricci curvature. The Ricci curvature of an edge in a graph describes the connectivity in the local neighborhood. An edge in a densely connected neighborhood has positive curvature and an edge serving as a local bridge has negative curvature. We use the edge curvature distribution to form a graph kernel which is then used to compare and cluster graphs. The curvature kernel uses purely the graph topology and thereby works for settings when node attributes are not available.  
The computation of the curvature for an edge uses only information within two hops from the edge and a random sample of $O(1/{\eps^{2}}\log 1/{\eps}+1/{\eps^{2}}\log 1/{\delta})$ edges in a large graph can produce a good approximation to the curvature distribution with error bounded by $\eps$ with probability at least $1-\delta$. Thus, one can compute the graph kernel for really large graphs that some other graph kernels cannot handle. This Ricci curvature kernel is extensively tested on graphs generated by different generative models as well as standard benchmark datasets from bioinformatics and Internet AS network topologies.

Graph classification and comparison are widely applied in bioinformatics, vision and social network analysis. 
One of the most popular approaches in practice is using graph kernels which compute the similarity of two graphs in terms of subgraph structures. Many graph kernels have been developed, which differ by the subgraph structures they focus on, such as random walks~\cite{gaertner03graphkernels}, shortest paths~\cite{borgwardt2005shortest}, 
 subtrees~\cite{ramon03graphkernels}, 
 and cycles~\cite{horvath04cyclicpattern}. Graph kernels have been extensively tested on benchmark datasets from bioinformatics to chemistry~\cite{borgwardt2005shortest,Ralaivola05}.
 
In our work, we focus on the setting of unlabeled graphs and propose a new graph kernel based on discrete Ricci curvature which takes only the network topology as an input. Our work is motivated by the use of curvature related kernels in shape matching.  Curvature on a smooth surface defines the amount by which a geometric object deviates from being flat or straight. It is a local measure at each point but nevertheless has deep connections to global topology and structures. Despite the success in shape matching, curvature has not been used much for comparing graphs. 
In this paper, we propose to use curvature distribution of graphs to build new graph kernels. The goal is to demonstrate that the Ricci curvature distribution and kernels can be efficiently computed and capture interesting graph properties. They add to the family of graph features and kernels, could be combined with other attributes, and used for other classifiers. 

\begin{figure}[tbp]
    \centering
    \includegraphics[width=\columnwidth]{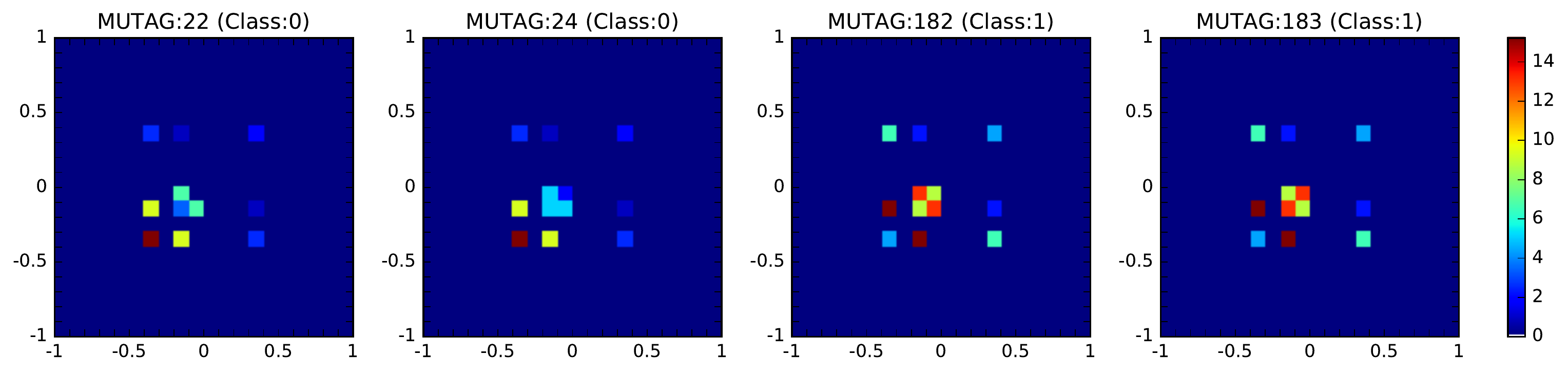}
    \caption{The 2D Ricci Curvature histogram of MUTAG graphs which describes the curvature distribution for pairs of neighboring edges. A suitable choice of mass distribution on the neighborhood results in curvature always ranging between $[-1,1]$. Graphs in the same class tend to have similar histogram distributions. Here MUTAG:22 and MUTAG:24 belong to the same class, while MUTAG:182 and MUTAG:183  belong to another class.}
    \label{fig:mutag}
\end{figure}

\section{Discrete Ricci Curvature}
For an edge $\overline{uv}$, define a distribution $m_u,m_v$ on the neighborhood of $u,v$ respectively (such as uniform $m_u$ and $m_v$). Now compute the Earth Mover Distance $W(u, v)$ from $m_u$ to $m_v$, where the cost of moving mass from a neighbor $u_i$ of $u$ to a neighbor $v_{j}$ of $v$ is the shortest path distance in the graph. Here the edges are unweighted unless they inherit weights arising from the application domain, such as tie strength or distances. For example, $W(u, v)$ will be upper bounded by 2 for unweighted graph if we allocate 50\% mass to the node's neighbor. The Ollivier-Ricci curvature is defined as 
$w(\overline{uv})=1-W(u, v)/d(u, v)$, where $d(u, v)$ is the length of edge $\overline{uv}$. Intuitively the curvature captures the structural properties of the local neighborhood. If $\overline{uv}$ stays in a well connected, dense neighborhood, the curvature is positive; if $\overline{uv}$ is locally a bridge, its curvature is negative. 

\section{Ricci Curvature Graph Kernel}
We define the Ricci Curvature Kernel as the following. Denote the curvature distribution of all edges in $G$ by $\textit{D}(G)$ and that of $G'$ by $\textit{D}(G')$. We use the standard Gaussian RBF kernel: $k(G, G')=\exp({-||\textit{D}(G)-\textit{D}(G')||^{2}_{2}/2\sigma^2}),$ where $||\textit{D}(G)-\textit{D}(G')||_{2}$ is the $\ell_2$ norm of two vectors $\textit{D}(G)$, $\textit{D}(G')$. Since the kernel depends on the curvature distribution, the distribution is less robust statistically for small graphs. We could boost up the kernel by considering the curvature distribution for pairs of neighboring edges $\{(w(e), w(e'))\}$. It appears to be more effective in practice. See Figure~\ref{fig:mutag} for an example. When the graph is really large, computing the curvature distribution might be costly ($O(|G.E|*n^3)$ where $n$ is the size of concerned neighborhood). Random sampling can be used to approximate the curvature distribution. Taking $O(1/{\eps^{2}}\log 1/{\eps}+1/{\eps^{2}}\log 1/{\delta})$ edges uniformly at random from the graph $G$, it can be shown that $\hat{\textit{D}}(G)$, the curvature distribution on the sampled edges, is a good approximation of $\textit{D}(G)$ with error bound $\eps$ with probability $1-\delta$. 
Notice that the running time does not even depend on the size of the graph $n$. 

\newpage
\bibliographystyle{splncs04}
\bibliography{reference_details}

\begin{thebibliography}{1}
\providecommand{\url}[1]{\texttt{#1}}
\providecommand{\urlprefix}{URL }
\providecommand{\doi}[1]{https://doi.org/#1}

\bibitem{borgwardt2005shortest}
Borgwardt, K.M., Kriegel, H.P.: Shortest-path kernels on graphs. In:
  Proceedings of the Fifth IEEE International Conference on Data Mining (ICDM
  2005). pp. 74--81. IEEE Computer Society, Washington, DC, USA (2005),
  \url{http://dx.doi.org/10.1109/ICDM.2005.132}

\bibitem{gaertner03graphkernels}
G{\"a}rtner, T., Flach, P., Wrobel, S.: On graph kernels: Hardness results and
  efficient alternatives. In: Schölkopf, B., Warmuth, M.K. (eds.)
  Computational Learning Theory and Kernel Machines --- Proceedings of the 16th
  Annual Conference on Computational Learning Theory and 7th Kernel Workshop
  (COLT/Kernel 2003) August 24-27, 2003, Washington, DC, USA. Lecture Notes in
  Computer Science, vol.~2777, pp. 129--143. Springer, Berlin--Heidelberg,
  Germany (August 2003)

\bibitem{horvath04cyclicpattern}
Horv{\'a}th, T., G{\"a}rtner, T., Wrobel, S.: Cyclic pattern kernels for
  predictive graph mining. In: Kim, W., Kohavi, R., Gehrke, J., DuMouchel, W.
  (eds.) Proceedings of the 10t ACM SIGKDD International Conference on
  Knowledge Discovery and Data Mining (KDD 2004), August 22-25, 2004, Seattle,
  WA, USA. pp. 158--167. ACM Press, New York, NY, USA (2004),
  \url{http://doi.acm.org/10.1145/1014052.1014072}

\bibitem{Ralaivola05}
Ralaivola, L., Swamidass, S.J., Saigo, H., Baldi, P.: Graph kernels for
  chemical informatics. Neural Netw.  \textbf{18}(8),  1093--1110 (2005),
  \url{http://dx.doi.org/10.1016/j.neunet.2005.07.009}

\bibitem{ramon03graphkernels}
Ramon, J., G\"artner, T.: Expressivity versus efficiency of graph kernels. In:
  Raedt, L.D., Washio, T. (eds.) Proceedings of the First International
  Workshop on Mining Graphs, Trees and Sequences (MGTS 2003) at the 14th
  European Conference on Machine Learning and 7th European Conference on
  Principles and Practice of Knowledge Discovery in Databases (ECML/PKDD 2003),
  September 22 and 23, 2003, Cavtat-Dubrovnik, Croatia. pp. 65--74 (2003)

\end{thebibliography}


\end{document}